%% file: main.tex
\documentclass[sigconf]{acmart}

\usepackage{booktabs} 
\usepackage{natbib}
\usepackage{amsfonts}
\usepackage{mathtools}
\usepackage{amsmath}

\DeclarePairedDelimiterX{\infdivx}[2]{(}{)}{%
  #1\;\delimsize|\delimsize|\;#2%
}

\setcopyright{acmcopyright}

\copyrightyear{2022}
\acmYear{2022}
\setcopyright{rightsretained}
\acmConference[SAC '22]{The 37th ACM/SIGAPP Symposium on Applied Computing}{April 25--29, 2022}{Virtual Event, }
\acmBooktitle{The 37th ACM/SIGAPP Symposium on Applied Computing (SAC '22), April 25--29, 2022, Virtual Event, }\acmDOI{10.1145/3477314.3507182}
\acmISBN{978-1-4503-8713-2/22/04}

\acmArticle{4}
\acmPrice{15.00}

\begin{document}
\title{Curriculum Learning for Safe Mapless Navigation}

  
\renewcommand{\shorttitle}{}

\author{Luca Marzari, Davide Corsi, Enrico Marchesini and Alessandro Farinelli}
\affiliation{%
  \institution{Computer Science Department, University of Verona}
  \streetaddress{Str. le Grazie, 15}
  \city{Verona} 
 \state{Italy} 
  \postcode{37134}
}
\email{luca.marzari@studenti.univr.it}


\input{Sections/abstract}


\maketitle

\input{Sections/introduction}

\input{Sections/preliminaries}
\input{Sections/methods}
\input{Sections/experimental_results}
\input{Sections/related_work}
\input{Sections/conclusion}

\begin{acks}
The research has been partially supported by the projects ”Dipartimenti di Eccellenza 2018-2022”, funded by the Italian Ministry of Education, Universities and Research(MIUR).
\end{acks}

\bibliographystyle{ACM-Reference-Format}
\bibliography{bibliography} 

\end{document}

%% file: Sections/abstract.tex
\begin{abstract}
This work investigates the effects of Curriculum Learning (CL)-based approaches on the agent's performance. In particular, we focus on the safety aspect of robotic mapless navigation, comparing over a standard end-to-end (E2E) training strategy. To this end, we present a CL approach that leverages Transfer of Learning (ToL) and fine-tuning in a Unity-based simulation with the Robotnik Kairos as a robotic agent. For a fair comparison, our evaluation considers an equal computational demand for every learning approach (i.e., the same number of interactions and difficulty of the environments) and confirms that our CL-based method that uses ToL outperforms the E2E methodology. In particular, we improve the average success rate and the safety of the trained policy, resulting in 10\% fewer collisions in unseen testing scenarios. To further confirm these results, we employ a formal verification tool to quantify the number of correct behaviors of Reinforcement Learning policies over desired specifications.
\end{abstract}

%% file: Sections/introduction.tex
\section{Introduction}

In recent years, Deep Reinforcement Learning (DRL) obtained impressive performance in a variety of fields such as games~\cite{silver2018general}, videogames~\cite{mnih2013playing}, robotic manipulation tasks~\cite{marzari2021towards} \cite{marchesini2019double}, and especially autonomous navigation in mapless context~\cite{8202134}. A key benefit for DRL methods with respect to the engineered solutions is the ability to adapt to unforeseen situations, and a crucial aspect of achieving this is the ability to generalize to scenarios that have not been used during training. A second crucial aspect to deploy DRL approaches to robotic platforms is related to safety. In particular, for mapless navigation, it is necessary to ensure that the robot actions do not cause damage to the robot itself and to the surrounding environment (i.e., objects and humans).\\
In this paper, we focus on the safety aspect of mapless navigation ~\cite{zhelo2018curiosity, marchesini2020genetic} using Deep Reinforcement Learning. In detail, our goal is to investigate novel training strategies that aim at improving safety, i.e., strategies that propose a solution to lower the number of collisions with obstacles during navigation both in the training and testing phases in previously unseen environments.\\
To this end, we leverage the idea of how humans exploit prior knowledge and learn in contexts they have never seen before. For example, humans and animals are prone to learn better behaviors when meaningful experiences are not randomly encountered but are organized in a meaningful order which presents a new concept in increasing order of complexity~\cite{curriculum}. Following this direction, our idea is to decompose a complex environment into (so-called) elementary components that are ordered based on a difficulty factor (i.e., with an increasing number of obstacles that reduce the available space for navigation). Our conjecture is that this helps in learning effective navigation and obstacle-avoidance skills, increasing the safety level of the resulting policy. In more detail, we hypothesize that an informed training strategy based on prior knowledge could lead to better performance over a naive and uninformed end-to-end (E2E) approach.\\
We present a comparative evaluation, using a simulated mobile robot for mapless navigation, between the standard E2E approach and our methodology, which is based on Curriculum Learning (CL)~\cite{curriculum}. In particular, the optimization process for our CL method is gradually scaled to approach the learning of the desired complex navigation behaviors. To this end, we employ techniques of Transfer of Learning (ToL)~\cite{transfer_2, parisotto2016actormimic} and fine-tuning, where the goal is to use or adapt the knowledge learned in a previous task to different application domains. 
For both the methodologies, we base our solution on the state-of-the-art Proximal Policy Optimization (PPO)~\cite{ppo} algorithm. \\
Our extensive evaluation in different environments aims to demonstrate that our CL based on ToL improves the robot's performance during the training (and testing) phase, increasing the safety, i.e., lowering the number of collisions in previously unseen environments.\\
We use the Robotnik Kairos mobile robot\footnote{https://robotnik.eu/products/mobile-manipulators/rb-kairos/} as an agent and create the training and testing environments using Unity\footnote{https://unity.com/our-company}. Specifically, we rely on the Unity ML-agents\footnote{https://github.com/Unity-Technologies/ml-agents} toolkit to ensure the communication between our training algorithms and the environments (hence, the agent).
To further support our claims on the benefits of CL, after the empirical results, we perform a formal verification of the networks trained with the different methodologies. For this purpose, we employ the state of the art tool for the formal analysis of neural networks in reinforcement learning scenarios, ProVe~\cite{prove}.\\
In summary, this work provides the following contributions to state-of-the-art\footnote{The source code of this work with all the results is free available at \url{https://github.com/LM095/Curriculum-Learning-for-Safe-Mapless-Navigation}}:
(i) we present a comparative study between CL-based methods combined with ToL or fine-tuning techniques and a standard E2E approach in a mapless navigation context to show that a CL-based approach improves the agent performance. (ii) we propose the use of a formal analysis tool to evaluate the safety of trained models to confirm our hypothesis of increasing the safety level by training a network in an informed fashion compared with an uninformed one. \\
Our results, obtained at equal computational effort (e.g., number of steps and difficulty of the environments) for all tested methodologies, confirm that our CL-based method that uses ToL outperforms the standard E2E approach in terms of average success rate and safety on unseen scenarios.  In particular, the approach proposed in this paper allows to increase the average safety level by more than 10\% compared to a standard E2E approach.

%% file: Sections/preliminaries.tex
\section{Preliminaries}
A popular model for Reinforcement Learning (RL) agent is Markov Decision Process (MDP). A MDP is represented by a tuple ($\mathcal{S}$, $\mathcal{A}$, $\mathcal{R}$, $\mathcal{P}$) \cite{Sutton2018} where $\mathcal{S}$ is the set of states, $\mathcal{A}$ is the actions space, $\mathcal{R}$ is a function defined as $\mathcal{R}\colon \mathcal{S} \times \mathcal{A} \to \mathbb{R}$ that, given the pair ($s \in \mathcal{S}, a \in \mathcal{A})$ returns a reward signal. Finally, $\mathcal{P}$ is the transition function $\mathcal{P}\colon \mathcal{S} \times \mathcal{A} \times \mathcal{S} \to [0, 1]$ that returns the conditional probability $\mathcal{P}(s' \mid s, a)$, where $s$ is the current state, $a$ is the action and $s'$ is the next state. 
The objective of an RL algorithm is to find a policy $\pi_\theta$ that maximize cumulative reward, discounted by a factor $\gamma$, formally denoted as:
\begin{center}
\begin{equation}
     \max_{\theta} \quad J(\pi_\theta) := \mathbb{E}_{\tau \sim \pi_\theta} [\sum_{t=0}^\infty \gamma^t R(s_t, a_t)]
\end{equation}
\end{center}
When the state space becomes not manageable for the traditional RL algorithm (e.g., continuous actions spaces, $\mathcal{S} \in \mathbb{R}^{n}$), a possible approach is to use Artificial Neural Networks (ANN) to represent the policy. In this context, the objective function (or loss function) can be expressed as follows \cite{Reinforce}:
\begin{center}
\begin{equation}
     \nabla J(\pi_\theta) := \mathbb{E}_{\tau \sim \pi_\theta} [\nabla_\theta \log_{\pi_\theta}(a_t \mid s_t)\mathcal{A}]
\end{equation}
\end{center}
However, with this formulation, the computed gradient suffers from high variance.  To address this issue, a standard approach is to use the advantage ($\mathcal{Q}$) instead of the standard episode reward. This value represents how good is this action with respect to the average of all the actions in the given state. Formally, the advantage can be expressed in different ways, a standard approach is $\mathcal{Q}(s, a) = \mathcal{R}(s, a) - \mathcal{V}(s)$. Where $\mathcal{V}(s)$ is a value function that can represent the mean of the reward obtainable from the state $s$ or, in an Actor-Critic configuration  \cite{actor_critic}, the value predicted by another neural network.
\subsection*{Proximal Policy Optimization}
To further improve the performance of the training, Trust Region Policy Optimization (TRPO) \cite{trpo} introduced the concept of \textit{trust region search strategy} to make sure that the new policy is not far away from the current policy, limiting the exploration only to the safe zone. 
To reach this objective, authors propose to use the \textit{KL-Divergence} to measure the distance between the current policy ($\pi_{\theta_{old}}$) and the new policy ($\pi_{\theta}$) and ensure that this distance is always less than a given value $\sigma$. Formally we rewrite the objective function as follows: 
\begin{equation*}
\begin{split}
    \max_{\theta} \quad J(\pi_\theta) &:= \mathbb{E}_{\tau \sim \pi_\theta} \big[\sum_{t=0}^\infty \gamma^t R(s_t, a_t)\big] \textrm{s.t.} \\ &D_{KL}(\pi_{\theta_{old}}||{\pi_{\theta}}) < \sigma
\end{split}
\end{equation*}

However, TRPO is a complicated algorithm that introduces a second-order optimization and adds additional overhead in the form of hard constraints. To improve this method to a first-order optimization problem, Schulman et al. propose Proximal Policy Optimization\cite{ppo}, authors encode the probability ratio of the new policy and old policy as follow:
\begin{center}
\begin{equation*}
\mathcal{r}(\theta) = \frac{\pi_\theta(a \mid s)}{\pi_{\theta_{old}}(a \mid s)}
\end{equation*}
\end{center}
Given $r(\theta)$ it is possible to rewrite the objective function as follows:
\begin{center}
\begin{equation*}
      J(\pi_\theta)^{clip} := \mathbb{E}_{\tau \sim \pi} [min(r(\theta)\mathcal{A}, clip(r(\theta), 1-\epsilon, 1+\epsilon)\mathcal{A}]
\end{equation*}
\end{center}
In the above equation, the ratio between the policies is always clipped in small intervals around 1. The size of the interval indicates the maximum allowed changes for the new policy, regulated by a hyperparameter $\epsilon$ (that was set to 0.2 in the original paper \cite{ppo}).
\subsection*{Curriculum Learning}
Curriculum Learning is initially proposed by Bengio et al. to improve the performance of machine learning models. In the original implementation, \cite{curriculum} CL is presented as a new paradigm for the training phase that gradually increases the complexity of the task (or data sample) during the process. Considerable research has been done to improve this concept in the last few years. A series of methods propose to increase the modeling capacity of the model \cite{curriculum_1} (e.g., adding neural units in an ANN model) or train the model with a process of activation/deactivation of the units \cite{curriculum_1}. In general, however, CL can be viewed as a continuation method \cite{curriculum_survey}, where the optimization process starts from a simplified version of the objective and is gradually scaled to the original and more complex function. 

This process is slightly different from the Transfer of Learning \cite{transfer}, where a model trained for one task is used as a starting point for the learning process of a new, although related, problem. The objective of ToL is to create models that have the ability to transfer knowledge across different tasks to improve their ability to generalize to new situations \cite{transfer_2}. A standard implementation in deep learning is to freeze some layer (or neural unit) of the ANN trained in the first loop before performing the next training phase.
Another common technique for ToL is Fine-Tuning. This family of approaches uses the weights of a previously trained ANN for a new training process. A possible implementation \cite{fine_tuning} is to use the same network trained for the first task inside the loop for the second one.
\subsection*{Formal Verification}
In the last year, DRL has been applied in several safety-critical domains, such as robotics \cite{robotics}, healthcare \cite{healthcare}, and navigation \cite{acas_xu}. In literature, a lot of methods have been proposed to improve the performance of the training from a safety point of view \cite{srl_survey}. However, the evaluation of safety in these contexts is an open problem.
Typically, ad-hoc metrics are designed for specific tasks. In autonomous driving and mapless navigation, for example, a standard option is to measure the average number of collisions in a set of sequential episodes \cite{SUPERL}. In the last year, the research of more formal methods for evaluating safety performance has been growing. 
However, Formal verification of neural network is still a hard problem, since the input space is continuous (i.e., $S \in \mathbb{R}^n$), it is not possible to test all the possible input configurations. Standard approaches \cite{verification_survey}, aims at verifying a given set of safety properties in the following form:
\begin{center}
\begin{equation}
\Theta: \mbox{If } x_0\in[a_0, b_0] \land ... \land x_n\in[a_n, b_n] \Rightarrow y_j\in[c, d]
\end{equation}
\end{center}
\noindent where $x_k \in X$, with $k \in [0, n]$ and $y_j$ is a generic output.
To of the first attempts to solve this problem are NeVer \citep{never}, which propose to abstract the verification problem with the interval algebra \cite{interval_algebra}, and ILP \cite{ILP} that, in contrast, encodes the verification problem as an optimization problem to solve with a linear programming approach. These first methods suffer from scalability problem that has been partially addressed by ReluVal \citep{reluval}, which proposes a \textit{symbolic propagation} approach to reduce the overestimation, and \cite{neurify} that introduce the concept of \textit{linear relaxation} to handle different types of nonlinear activation functions. However, in our work, we use ProVe \cite{prove} for the evaluation, an evolution of the previous methods specifically designed for an RL scenario. In particular, they encode the properties in a different fashion:
\begin{center}
\begin{equation} 
\Theta: \mbox{If } x_0\in[a_0, b_0] \land ... \land x_n\in[a_n, b_n] \Rightarrow y_j > y_i
\label{eq:prove_property}
\end{equation} 
\end{center}
Moreover, they proposed a metric to evaluate the safety based on a given set of properties, the \textit{violation rate}, defined as the percentage of the properties that cause a violation.

%% file: Sections/methods.tex
\section{Methods}
\label{sec:methods}
Our goal is to demonstrate that by using an approach based on CL combined with techniques of ToL or fine-tuning, we can achieve a higher success rate and, crucially, increase the safety of the policy in unseen environments. We now present the network architecture we use for all methodologies considered and our methods to train these networks. We present a formal verification of the trained models using these different methodologies in Section~\ref{exp:formalVerification}.
\subsection{Network Architecture}
\begin{figure}[h!]
    \centering
    \includegraphics[width=\linewidth]{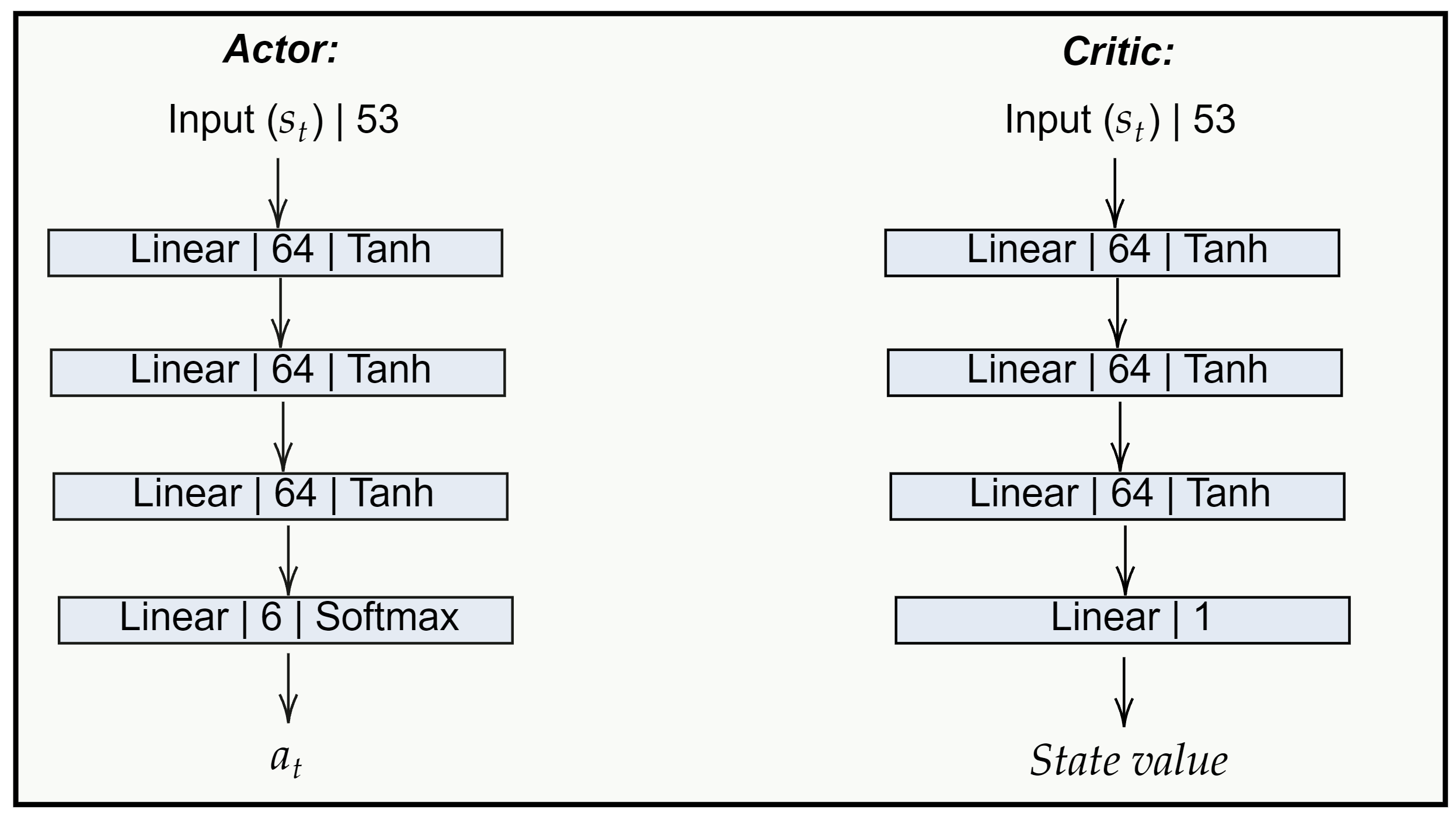}
    \caption{Network architecture for the proposed methodologies. Every layer is represented by type, dimension and activation function in PyTorch nomenclature}
    \label{fig:architecture}
\end{figure}

We consider a mapless navigation scenario, where a robot is positioned at a random point on the map without global knowledge of its surroundings, while it locally senses the environment with two 270-degree laser sensors placed on it combined with a 360-degree scan. In contrast to prior studies based on CL~\cite{chaffre2020sim}, our work uses random starting position and goal position spawn to favor generalization to unseen scenarios, hence avoiding overfitting to specific configurations of the environment. All the proposed methodologies share the same network architecture: an input layer consisting of 51-sparse 360-degree laser scans and two inputs for the target position, which is expressed in polar coordinates, and report the distance and angle (in degrees) with respect to our RB Kairos agent. These input values are normalized in the range $(0, 1)$. Concerning the number of hidden layers, we use three layers, each composed of 64 nodes with a Tanh activation. This configuration stems from the training methodology ToL adopted and presented in the section~\ref{ToL} and is in line with the typical architectures that we find in the literature for mapless navigation\cite{drl_navigation2}.\\
Finally, we use 6 output nodes with a Softmax activation, where each node encodes an action $a_t$ to be performed, i.e., every possible direction $(-90,-45,0,45,90)$ in addition to left and right rotation since the Kairos robot considered in this paper has three axes of freedom.
For all the methodologies, we use a discrete \textit{actor-critic} implementation of PPO. The decision to use this version of PPO stems as prior works have shown that discrete DRL approaches can achieve comparable performance over continuous one in the context of mapless navigation~\cite{drl_navigation2} while significantly improving sample efficiency. A complete schema of the architecture used in this work is depicted in Fig.~\ref{fig:architecture}\\
\\
Regarding the reward function, several tests have been performed to find a suitable reward function that incentives the robot to make the safest trajectory to avoid obstacles. In order to have a valid comparison, all the tested methodologies share the same reward function:
\begin{equation}\label{rewardFunc}
    r_t = \begin{cases}
            R_{crash}\quad \text{if crashes}\\
            R_{reach}\quad \text{if reaches the goal}\\
            \psi\cdot(d_{t-1} - d_t)\quad \text{otherwise}
          \end{cases}
\end{equation}
The reward function is thus half sparse and half dense. In particular, the values $R_{crash}$ and $R_{reach}$ are obtained when the robot hits an obstacle or when it reaches the target point, respectively, while in all other situations, the agent obtains a dense reward which is given by a factor $\psi$, multiplied by the difference in displacement $d$ between the previous timestep $t-1$  and the current timestep $t$.\\
Hence, we note that the reward can also be negative in case the robot does not make the direct trajectory towards the goal. To incentive the agent to do the safest trajectory, i.e., the trajectory that allows to lower the collision rate, we set the following parameters: $R_{crash}$ = -100, $R_{reach}$ = 300 and $\psi$ = 10, which resulted in better navigation performance in our evaluation. The values reported are those used in the experiments presented in Section \ref{results} and follow the typical shape of reward functions for mapless navigation in the literature~\cite{ray2019benchmarking}. 

\subsection{Training}\label{training}
This section presents the three training methodologies used in this work, namely Transfer of Learning, fine-tuning, and finally, the classical E2E training. Our idea is based on taking a complex environment, breaking it down into its basic components, and creating environments in order of difficulty to create increasingly informed knowledge about navigation. To this end, the first thing that needs to be defined is what we mean by difficult. In this work, we consider the area occupied by obstacles and the smallest distance between two obstacles in the environment itself as a metric of difficulty. \\
The biggest difficulties we find in a navigation task are very close obstacles, narrow corridors, \textit{U-shaped} walls. Hence, we insert all these complex components in a final environment, which we refer to as \textit{finalEnv} in Fig.\ref{fig:envs}, and then train the agent with three different methodologies. We propose two methodologies that aim to create an increasingly informed knowledge about the task, respectively using ToL and fine-tuning, and one directly in an uninformed fashion,i.e., directly training the agent in the \textit{finalEnv}, using E2E.
\subsubsection*{Transfer of Learning}\label{ToL}
\begin{figure}[h!]
    \centering
    \includegraphics[width=\linewidth]{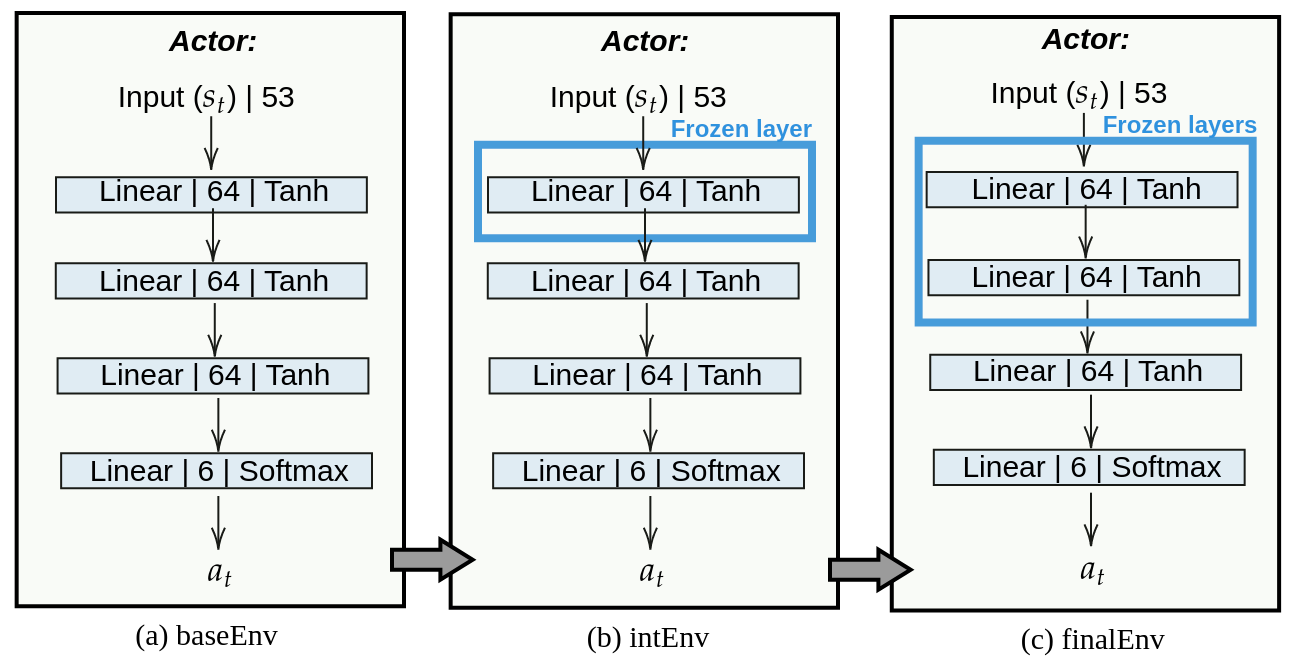}
    \caption{Schematic approach used for training through Transfer of Learning (ToL)}
    \label{fig:transfer}
    \vspace{-4mm}
\end{figure}

Transfer of Learning, or simply Transfer Learning, is a machine learning technique, where the knowledge gained during training in one type of problem is used to train in other related tasks or domain~\cite{transfer_2}. This technique is generally used for image classification using Convolutional Neural Network (CNN), but here we use it instead in the context of DRL.\\
In order to build an increasingly informed knowledge about navigation, we decide to start from the basis of navigation, i.e., an empty room with only two obstacles placed inside it. In this environment, called \textit{baseEnv}, the robot learns to navigate and reach the goal efficiently in terms of the number of steps, i.e., without superfluous actions and begins to understand that it receives a negative reward if it hits an obstacle. Once we reach a high success rate in this room, we load the next environment dynamically. Kindly note that we dynamically load the environment here, and we do not interrupt the training. 
So after loading the new environment and before starting the new training, we freeze the first of the three hidden layers. When we talk about freezing a layer, we mean that the weights of that layer are considered constant from that moment on, and no backpropagation on that layer is performed during the network update.\\
In this new environment, called \textit{intEnv}, which stands for intermediate environment, we try to make the robot understand that a longer trajectory that avoids the obstacles is the desired behavior. In particular, inside this room, we insert a \textit{U-shaped} wall, which is particularly challenging for navigation. At the end of the training in this room, we proceed similarly in the complex environment that we divided initially: so we dynamically load the environment called \textit{finalEnv} and freeze the first two layers, leaving only the last one trainable. What basically happens is a direct mapping between environment and layer, the knowledge acquired on the first environment is maintained thanks to the freezing of the first layer, while the knowledge of the second environment with the freezing of the first two. With all the information acquired on these first two environments, we start the training on the final environment.\\
In this work, we have considered only two environments before the final environment, and from the results presented in Section~\ref{results}, they seem to be sufficient to obtain a good level of generality and safety in different unseen environments. A schematic approach used for training through ToL is presented in Fig~\ref{fig:transfer}.
\subsubsection*{Fine-tuning}
Fine-tuning is a concept very close to ToL. To investigate how much the freezing action impacts the agent's performance using an informed technique, we introduce this second informed training method called fine-tuning. This methodology consists of training the agent analogously to what is done in ToL, but in this case, without freezing the layers from one environment to another. In other words, we start from the first \textit{baseEnv} environment with the same characteristics described above, and once we reach a high success rate, we dynamically switch to the second \textit{intEnv} environment. Here, we do not freeze the weights of the first layer, but we start directly with the training. Similarly, we perform with the final environment. In this way, after the first backpropagation, the weights obtained from the training on the first environment are modified, allowing a comparison between the two methodologies, i.e., we can understand how much influence it has or not to freeze the layers during the informed training. 
\begin{figure*}[h!]
    \centering
    \includegraphics[width=\linewidth]{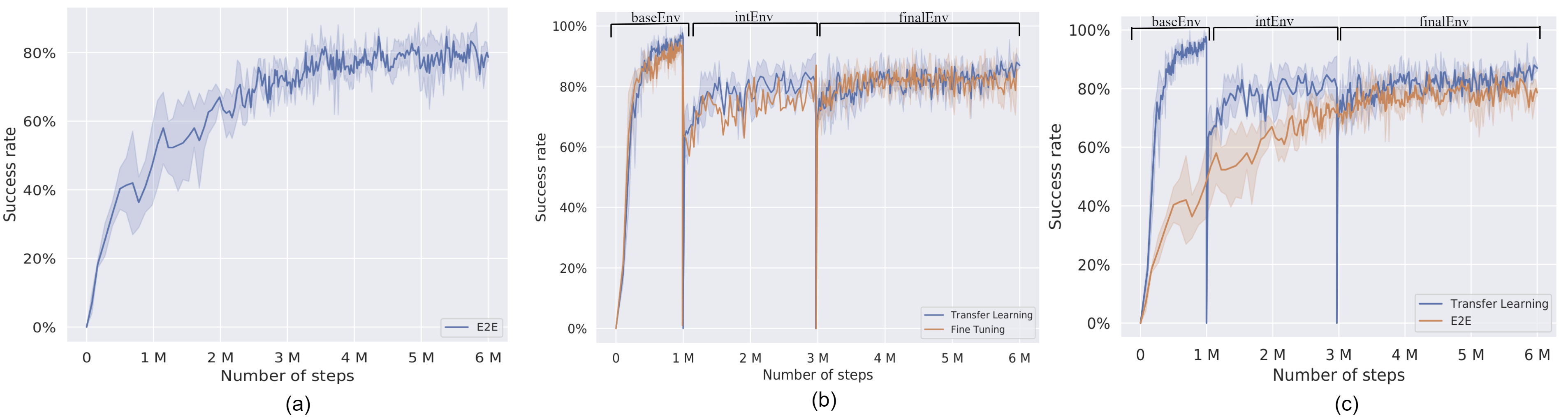}
    \caption{(a) E2E plot. (b) Plot comparison of ToF and Fine-tuning. The value reported is the median of the best performance (success rate) of all randomly seeded runs of each method. The first peak, for the informed training, denotes the completion of training in the \textit{baseEnv}, the second peak denotes the completion of training in the \textit{intEnv}, and similarly, the third peak denotes the completion in the \textit{finalEnv}. (c) Comparison between E2E and ToL: starting and goal positions are randomly chosen, so they may not coincide at the same timestep.}
    \label{fig:plotTraining}
\end{figure*}
\subsubsection*{End-to-End}
The E2E training involves classical uninformed training, using our PPO implementation directly on the \textit{finalEnv} environment. To ensure a fair comparison, we first start with this approach to know how many timesteps are required to reach a high success rate in the \textit{finalEnv} environment, and then we divide the timesteps employed by E2E in the respective timesteps for each sub-environment of the informed strategies described above. Section~\ref{results} presents in detail all the experiments performed to compare the methodologies described here.

%% file: Sections/experimental_results.tex
\section{Experimental Results}\label{results}
In order to validate our hypotheses, we perform two sets of experiments: a first set of experiments comparing informed and uninformed training methodologies during the training phase and a second set concerning the testing phase in different unseen environments, where we test the level of adaptability, safety, and skill acquired during the previous training phase.\\
Given that we trained the robot using random spawn of both initial position and goal position, we performed this evaluation over several episodes. In particular, we report the average results obtained by testing the agent on three different randomly seeded runs of 1000 episodes each.\\
All tests are performed on an Intel Core i5-11600K, using, as mentioned above, Adam~\cite{kingma2014adam} as an optimizer with an initial learning rate of 0.0003. During the training phase, we set $\gamma = 0.99$ and the $\epsilon$-clipping $= 0.2$ as suggested in the original paper of PPO~\cite{ppo} and we update the network every 6000 steps.
\subsection{Training experiment}
\begin{figure}[h!]
    \centering
    \includegraphics[width=\linewidth]{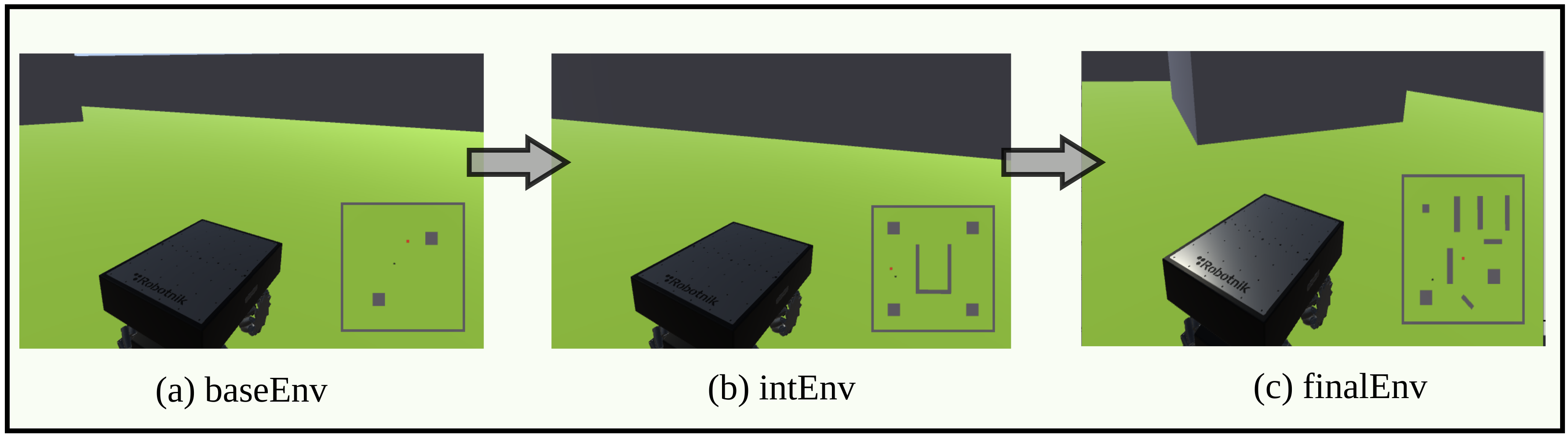}
    \caption{Environments created in Unity and used for the training phase}
    \label{fig:envs}
    \vspace{-5mm}
\end{figure}

We create a final environment, called \textit{finalEnv} (Fig. \ref{fig:envs}c), for the training test with the difficulties for navigation that we described in Section~\ref{training}. In particular, we try to reproduce narrow corridors and put obstacles in a position to force the robot to take the most extended trajectory to reach the goal as safely as possible.\\
Once we create this environment, we start with E2E training in an uninformed fashion to see how many steps are required to get a high success rate. In order to make a comparison between this training technique and the informed technique proposed in this work, we divide the final environment into its basic components, trying to create an increasing difficulty. Moreover, to ensure that each methodology trains on the same number of interactions, we split the E2E steps (required to get a high success rate) into smaller ranges for each sub-scenario. We depict in the Fig.~\ref{fig:envs} the environments created to perform the experiments in this phase.
\begin{figure*}[h!]
    \centering
    \includegraphics[width=\linewidth]{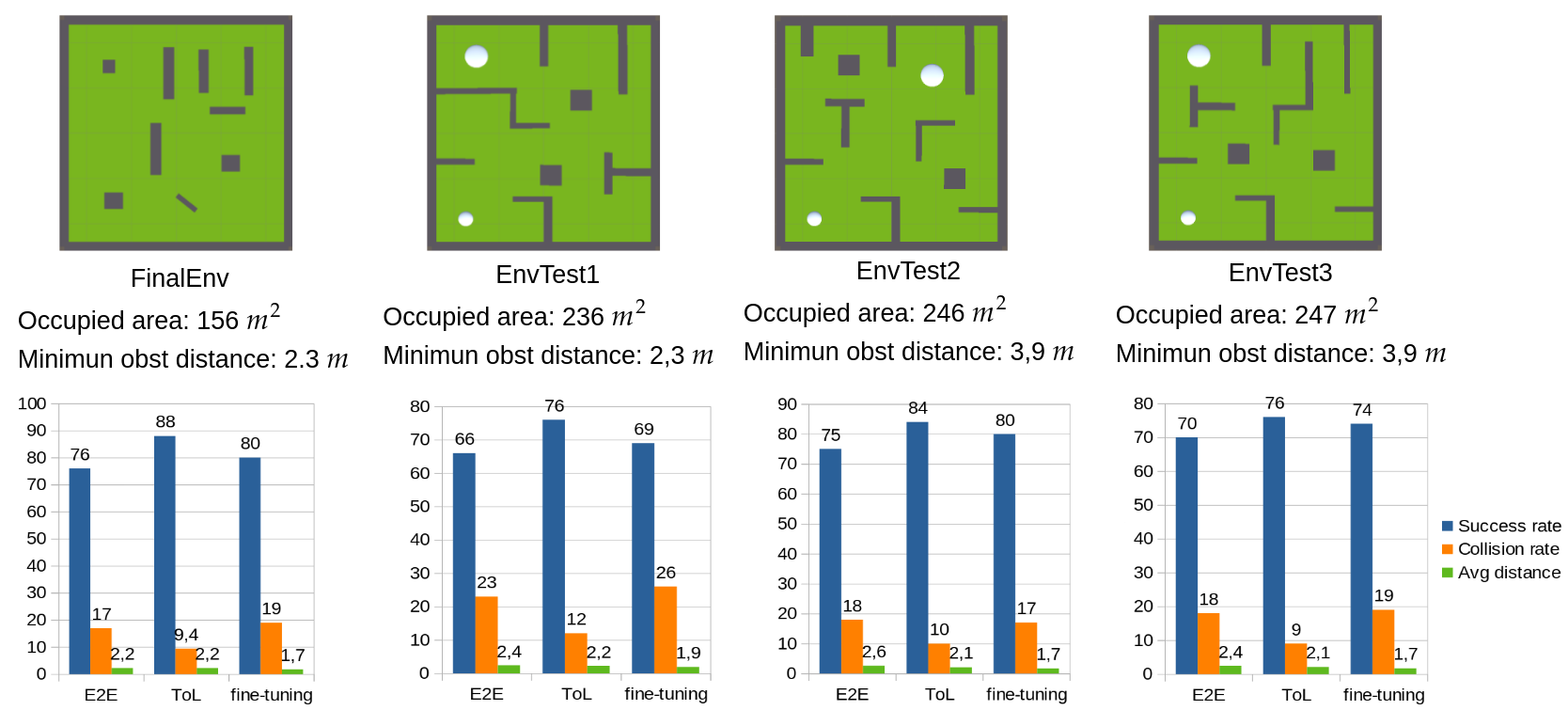}
    \label{fig:exp1}
\end{figure*}
\begin{figure}
    \centering
    \includegraphics[width=\linewidth]{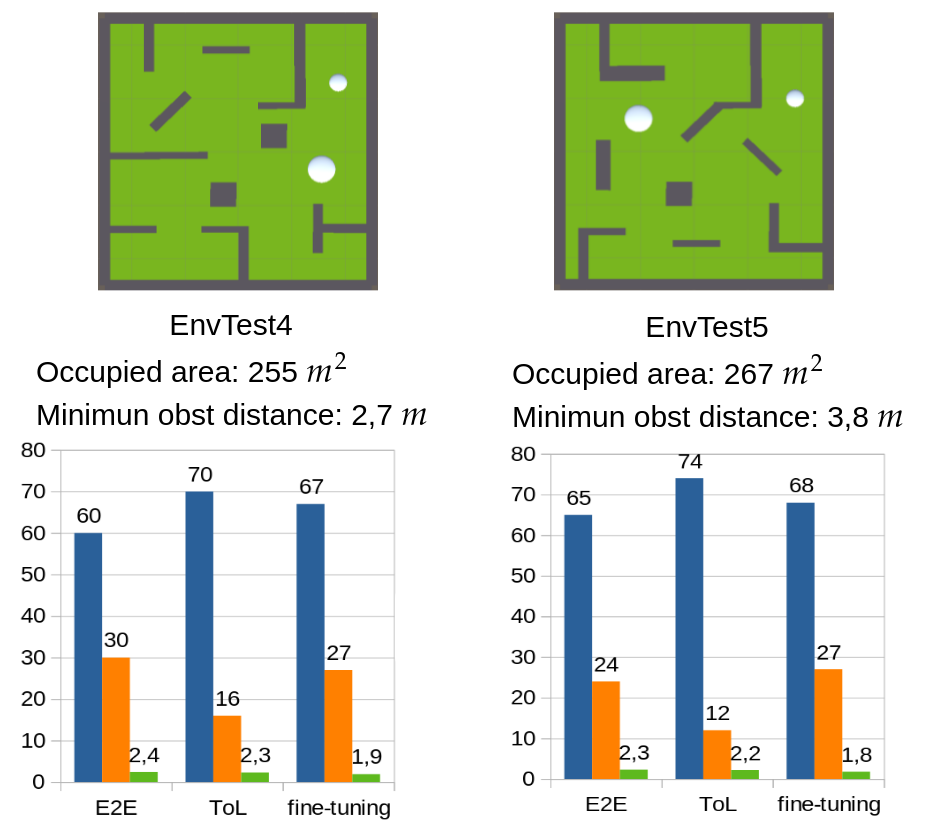}
    \caption{Results of the testing phase: for each environment, we report the area occupied, the minimum distance between two obstacles in the environment, and a bar graph showing average success, average collision rate, and the average deviation of the path.}
    \label{fig:exp2}
    \vspace{-5mm}
\end{figure}
Fig.~\ref{fig:plotTraining} depicts the success rate during the training phase for both uniformed and informed strategies. We can notice that the steps required to reach the constant high success rate by the E2E methodology (Fig.~\ref{fig:plotTraining}a) are 6M, so we divide the steps for the informed methodology into respectively 1M steps for \textit{baseEnv}, 2M for \textit{intEnv}, and finally 3M steps for \textit{finalEnv}. We can also see that both methodologies in the final environment succeed above 80\% during the training phase (Fig. \ref{fig:plotTraining}a and Fig.\ref{fig:plotTraining}b from in the \textit{finalEnv} part). ToL results to be the best strategy for the informed strategies, as it presents a smoother learning curve on each environment, in particular in the last steps of the final environment where fine-tuning is slightly unstable. Hence, this suggests that the freezing action of the layers presented in Sec.~\ref{ToL} actually allows preserving part of the knowledge that is lost over time otherwise.
Moreover, Fig.~\ref{fig:plotTraining}c depicts the comparison graph between uninformed and informed ToL methodology. In the first 3M steps, ToL acquires incremental knowledge about navigation in the \textit{baseEnv} (Fig. \ref{fig:envs}a) and \textit{intEnv} environment (Fig. \ref{fig:envs}b) respectively, and at the same time, the uninformed methodology (E2E) has not yet reached an average success rate above 80\% in the \textit{finalEnv}. We note in Fig.~\ref{fig:plotTraining}c at 3M steps how the information acquired in the first two scenarios allows the agent trained through informed techniques to immediately achieve a success that exceeds the one of E2E in an unseen scenario (\textit{finalEnv}). The difference, though initially minimal, emphasizes an advantage of these informed techniques: the greater degree of adaptability to unseen environments compared to classical uninformed training.\\
Another aspect that can also be noted is that in general, ToL has a higher average success than E2E, this is probably due to the knowledge acquired in particular on the second \textit{intEnv} environment, which allowed the agent to acquire more skills in particular on making longer but safer trajectories. 

\subsection{Testing experiments}
In the second set of experiments, i.e., those inherent to the test phase, we investigate the level of adaptability and skills acquired by the agent trained with the proposed various methodologies.\\
The test is performed on five new unseen environments besides the one where we trained the robot. These new scenarios all have similar characteristics to the \textit{finalEnv}, and to establish the difficulty level, we define two metrics: the area occupied by obstacles and the minimum distance between two obstacles in the environment. Each environment represents a squared room with dimensions $50m\times50m$, therefore with an area of $2500\;m^2$, that could represent a warehouse, inside which are positioned various walls and obstacles of different geometric shapes, perhaps also not seen during the phase of training.\\
On each environment and for each methodology were collected several metrics: the average success rate, the average collision rate, i.e., on $100$ episodes how many times an agent hits a wall or an obstacle, and the average deviation of the path taken by the agent compared to the direct trajectory, i.e., if the starting position is at a distance of $1$ from the goal (considering the obstacle) and the agent makes a distance $2$ means that it has made a distance twice as long as the direct trajectory. We note that this last metric indicates how much the robot has learned the trade-off between traveling a higher distance and avoiding collisions.\\
\begin{table*}[h!]
\begin{tabular}{p{0.15\linewidth}ccccccc}
\hline\noalign{\smallskip}
\textbf{Method} & \multicolumn{6}{c}{\textbf{Environments}} & {\textbf{Tot avg gain}} \\
\noalign{\smallskip }\cline{2-7} \noalign{\smallskip} 
    &        FinalEnv &          EnvTest1 &        EnvTest2 &          EnvTest3 &  EnvTest4 & EnvTest5  \\
\noalign{\smallskip}\hline\noalign{\smallskip}
ToL & 7.6\% & 11\% & 8\% & 9\% & 14\% & 12\% & \textbf{10.1}\%\\
Fine-tuning & -2\% & -3\% & 1\% & -1\% & 3\% & -3\% & \textbf{-0.8}\%\\
\noalign{\smallskip}\hline
\end{tabular}
\caption{comparison of average collision rate gain by ToL and fine-tuning with respect to the uninformed E2E methodology.}
\label{tab:tableAvgCollision}
\vspace{-5mm}
\end{table*}
Fig.~\ref{fig:exp2} depicts the results of our evaluation. In particular, we order the environments by the surface area occupied by obstacles. The results, in general, show that an agent trained with an informed strategy, in particular through the use of techniques such as ToL, can have a better generalization and adaptability to unseen environments achieving a higher success rate and safety level measured here as the number of collisions. In the testing phase, ToL is confirmed as the best-informed methodology, as we obtained a higher average success rate over the uninformed E2E methodology and the informed fine-tuning.
Moreover, the most important thing to point out is that the average collision rate with ToL is almost halved in each test environment compared to E2E.\\
Table~\ref{tab:tableAvgCollision} shows the average rate of gain inherent in safety using informed techniques such as ToL and fine-tuning compared to E2E. As we can see, ToL allows us to be on average 10\% safer, while fine-tuning resulted in $\approx 1\%$ lower safety score. 
Our hypothesis is that fine-tuning methods do not involve layer freezing, and therefore the various information acquired in one environment are only adapted to new situations encountered when changing the environment. In particular, this is reflected in the data regarding the average deviation from the direct trajectory. The agent trained by fine-tuning tends to take a slightly more direct trajectory than, for example, ToL that follows a longer trajectory but leads to manage to lower collision rate.
\subsection{Formal verification}\label{exp:formalVerification}
In safety-critical contexts, where human safety and expensive equipment could be involved, it is important to provide formal guarantees before deploying in a real-world environment. Our goal is to provide some safety guarantees that go beyond empirical evaluation. 
This section shows the results of our formal analysis of the trained models. For the process, we rely on ProVe \cite{prove}, a state-of-the-art tool for the formal analysis of neural networks in reinforcement learning scenarios. We design a set of 5 high-level properties, that describe the behavior of the agent to ensure that it makes rational decisions, following the \textit{behavioral properties} paradigm \cite{prove}, as follows:

\textbf{$\Theta_{0}$:} If there is an obstacle near to the left, whatever the target is, go straight or turn right.

\textbf{$\Theta_{1}$:} If there is an obstacle near to the right, whatever the target is, go straight or turn left.

\textbf{$\Theta_{2}$:} If there is an obstacle near to the left  and an obstacle in front, whatever the target is, turn right.

\textbf{$\Theta_{3}$:} If there is an obstacle near to the right and an obstacle in front, whatever the target is, turn left.

\textbf{$\Theta_{4}$:} If there are obstacles near both to the left and to the right, whatever the target is, go straight. 

The set of these properties describes, at a high level, safe behavior for the agent \cite{verification_survey}. To formally rewrite the properties, we recall the structure of the network shown in Section \ref{sec:methods}. We also consider an obstacle \textit{near} when the value of the corresponding input scan is less than $0.25$. For example, we rewrite $\Theta_{0}$ as follows:
$\Theta_{0}: \mbox{If } x_0, ..., x_8 \in [0, 1] \land x_9, ..., x_{18} \in \mathcal{D} \land x_{19}, ..., x_{50} \in [0, 1] \Rightarrow [y_3, y_4] < [y_0, y_1, y_2], where\; \mathcal{D} = [0, 0.25)$
\begin{table}[h!]
    \centering
    \begin{tabular}{lccc}
      \toprule
      \bfseries Property & \bfseries Transfer (ToL) & \bfseries End to End & \bfseries Fine-Tuning \\
      \midrule
      \textit{$\Theta_{0}$} & 5.7 $\pm$ 0.3  & 12.5 $\pm$ 1.3  & 11.4 $\pm$ 0.4 \\
      \textit{$\Theta_{1}$} & 6.3 $\pm$ 1.1  &  8.8 $\pm$ 0.6  &  9.4 $\pm$ 0.5 \\
      \textit{$\Theta_{2}$} & 5.2 $\pm$ 0.2  & 11.4 $\pm$ 0.2  &  8.5 $\pm$ 0.1 \\
      \textit{$\Theta_{3}$} & 9.6 $\pm$ 0.5  &  9.2 $\pm$ 0.5  &  8.4 $\pm$ 0.8 \\
      \textit{$\Theta_{4}$} & 7.5 $\pm$ 0.9  & 10.3 $\pm$ 0.6  &  9.4 $\pm$ 1.3 \\
      \midrule
      \textbf{Mean:} & \textbf{6.86} & \textbf{10.44} & \textbf{9.42}   \\
      \bottomrule
    \end{tabular}
    \caption{Comparison of the safety analysis results based on the \textit{violation rate} (\%) \cite{prove}, on our different training approaches: Transfer of Learning, End to End and Fine-Tuning.}
    \label{tab:resutls:prove}
    \vspace{-7mm}
\end{table}
Table \ref{tab:resutls:prove} shows our results. To analyze the statistical significance of the data \citep{colas2019hitchhikers}, we report the mean and standard deviation of the violation metric (\%), for each property and method, from the 5 best models and 3 seeds. The formal analysis of the trained network confirms our empirical results. Overall, Transfer of Learning results the safest approach, with the lowest violation rate, while the End to End training results in the highest violation rate. 

%% file: Sections/related_work.tex
\section{Related Work}

In the literature, a mapless navigation problem is typically modeled as a Markov Decision Process (MDP) \cite{8202134}. In more detail, robotic navigation has served as a widely adopted benchmark for robotics research \cite{drl_navigation1, drl_navigation2} due to its practical implications.

A line of research employs supervised learning approaches for mapless navigation. However, the non-trivial cardinality of labeled data required by such learning approaches is a significant limitation. A different perspective \cite{supervised_navigation1, supervised_navigation2} aims at using human intervention or demonstrations to collect such data from a realistic scenario. However, this research trend requires image processing, resulting in computationally demanding solutions \cite{supervised_navigation3}. 

In contrast, the trial-and-error nature of DRL makes it well-suited to decision-making tasks, such as navigation. Following this direction,
the typical trend in robotic research employs policy-gradient (or actor-critic) approaches, due to the intuitive way in which continuous action spaces can accurately model joint velocities in robotics \cite{8202134, drl_navigation3}. 
However, recent work \cite{drl_navigation2} showed the benefits of using value-based DRL in this domain, drastically reducing the computational complexity while maintaining comparable performance over policy-gradient (or actor-critic) solutions. 

Reality gap \cite{sim_to_real1} is also an important factor to consider when a policy learned in simulation is transferred to reality. However, most state-of-the-art results in DRL have been achieved using simulation and transferring the policy on real platforms \cite{unity_sim, rubik_hand, sim_to_real2, sim_to_real3}, showing promising results. Hence, commonly to prior DRL approaches for navigation \cite{drl_navigation2}, our ANN architecture relies on sparse laser scans and the goal coordinate to learn navigation skills in simulation.

%% file: Sections/conclusion.tex
\section{Conclusion}
In this paper, we investigate how much an informed training methodology combining Curriculum Learning and Transfer of Learning or fine-tuning techniques can be better than a classic End-to-End uninformed methodology in a mapless navigation context, with a particular focus on the safety aspect. Specifically, through our experiments, we demonstrated how much the freezing action of various layers in the informed solutions affected agent performance. Our results suggest that the informed solutions, particularly those that make use of the ToL, allow to reach a better success during the training and guarantee more safety in unseen environments with respect to the uninformed ones. The correctness of our conclusions was confirmed by a formal analysis of the trained models with different methodologies presented.\\
Regarding future work, we intend to investigate whether the results obtained remain consistent even with dynamic obstacles within the environment and on the transportability of this methodology in the real world scenarios.